# Soft Pneumatic Actuator Capable of Generating Various Bending and Extension Motions Inspired by an Elephant Trunk

Peizheng Yuan, Hideyuki Tsukagoshi

*Abstract*—Inspired by the dexterous handling ability of an elephant's trunk, we propose a pneumatic actuator that generates diverse bending and extension motions in a flexible arm. The actuator consists of two flexible tubes. Each flexible tube is restrained by a single string with variable length and tilt angle. Even if a single tube can perform only three simple types of motions (bending, extension, and helical), a variety of complex bending patterns can be created by arranging a pair of tubes in parallel and making the restraint variable. This performance takes advantage of the effect of the superposition of forces by arranging two tubes to constructively interfere with each other. This paper described six resulting pose patterns. First, the configuration and operating principle are described, and the fabrication method is explained. Next, two mathematical models and four finite element method-based analyses are introduced to predict the tip position changes in five motion patterns. All the models were validated through experiments. Finally, we experimentally demonstrated that the prototype SEMI-TRUNK can realize the action of grabbing a bottle and pouring water, verifying the effectiveness of the proposed method.

## I. Introduction

Soft pneumatic actuators (SPAs) for soft robots have rapidly evolved in recent decades [1]. Unlike traditional robots with rigid joints, SPAs are primarily made of soft materials like rubber [2]. Because of their intrinsic dexterity and compliance, SPAs can achieve a larger degree of freedom (DOF) with a simpler structure compared to their rigid-body counterparts [3]. This makes them ideal for performing complex tasks in the real world, such as exploring complicated environments [4].

To achieve innovative kinematic mechanisms and large DOFs, many SPAs have been designed to mimic biological systems [5]. By drawing inspiration from mammals, plants, insects, and other organisms, biomimetic SPAs with special hardware designs and actuation methods have been shown to be reliable for inspecting specific environments. For example, various inchworm-inspired SPAs have been applied in pipe inspection robots [6]-[8]. However, more complex deformations like turning require aligning several extensible SPAs in parallel [9] or using specific bending mechanisms [10]. Although bioinspired SPAs have been highly regarded for realizing specific application scenarios, the need for developing bio-inspired SPAs with multifunctionality remains urgent, given the complexity of exploring the real world [11].

To achieve multiple motion patterns, mimicking natural objects like an octopus arm or elephant trunk has become a popular approach in recent years [12]. The octopus arm and elephant trunk have similar anatomical structures, consisting entirely of soft tissue, with a central cord structure surrounded by transverse muscles and bundles of longitudinal muscle fibers arranged symmetrically [13]-[15].

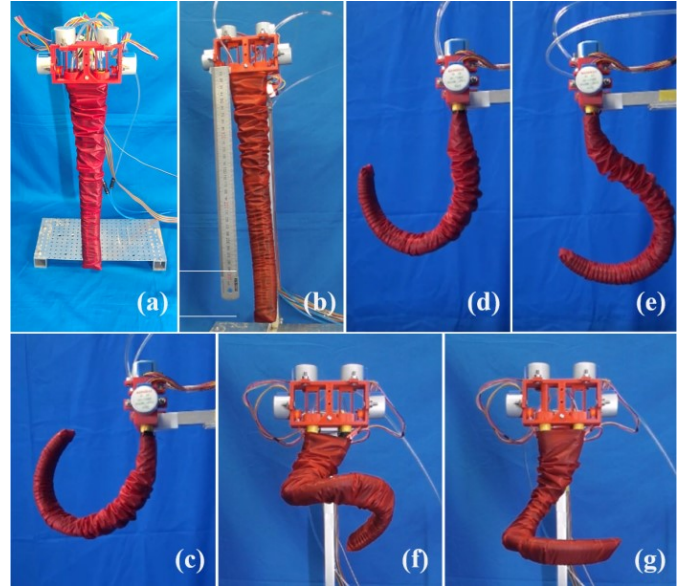

Figure 1. (a) The proposed SEMI-TRUNK can realize (b) linear extension, (c) C-shaped bending, (d) J-shaped bending, (e) S-shaped bending, (f) helical bending, and (g) spiral bending.

Some researchers have attempted to simplify the structural complexity of the octopus arm and elephant trunk when building their SPAs. For instance, Laschi developed a soft robot arm in 2012 that used four controllable longitudinal fibers to mimic all the longitudinal muscles in the octopus arm [16]. Another study in 2020 combined several cylindrical actuators serially with various functions, such as bending, twisting, contraction, and extension, to mimic the elephant trunk [17]. However, while these simplified structures may have small volumes, they often cannot realize the complex motions of the appendages they imitate. In contrast, some research efforts, such as the FESTO robot arm [18] and the fruit harvesting manipulator [19], have attempted to arrange a large number of contraction/extension actuators symmetrically to imitate the enormous muscle fibers in the elephant trunk, resulting in the ability to realize complex motion patterns. However, these structures are often bulky and heavy, making them difficult to use in real-world applications.

To address the tradeoff between functionality and compactness of elephant-trunk-inspired robots, our previous research has focused on developing the helical actuator [20] and the double helical actuator [21]. Now, we present the next generation of our work: SEMI-TRUNK, as shown in Fig. 1(a). This novel actuator can realize six motion patterns inspired by the elephant trunk in 1D/2D/3D spaces, as shown in Fig. 1(b)-

Peizheng Yuan is with Tokyo Institute of Technology, Tokyo 152-8550, Japan. (phone&fax: +81-3-5734-3085; e-mail: yuan.p.aa@m.titech.ac.jp; address: 2-12-1-S5-19, Ohokayama, Meguro-ku, Tokyo 152-8550, Japan).

Hideyuki Tsukagoshi is with Tokyo Institute of Technology, Tokyo 152-8550, Japan.

(g). It can be fabricated using simple and inexpensive materials, such as rubber tubes, textile watering hoses, threads, and stepper motors, enabling it to maintain a compact volume. Through the combination of different motion patterns, the SEMI-TRUNK can undertake complex tasks in the real world.

## II. Driving Principle

### A. Mechanical design.

To preserve the double helical actuator's deformability, the SEMI-TRUNK retains the basic framework of aligning two symmetrical tubular sub-actuators.

The mechanical structure of the SEMI-TRUNK is depicted in Fig. 2. Two rubber tubes are symmetrically arranged in an inverted trapezoid shape and connected at the bottom with a plastic frame. Textile hoses with zigzag surfaces are wrapped around the tubes to allow longitudinal extension but restrict transverse deformation. The tops of the tubes are connected to a custom frame that allows them to rotate. Two motors (a1, a2) are attached to the rotatable parts. Inextensible threads pass through the wedges formed by the zigzag surfaces of the hoses, with one end fixed on the bottom frame and the other connected to two additional motors (b1, b2). The threads are thin, smooth, longer than the tubes, and wound around the rotation shaft of the motors to enable low-friction control of the thread length that goes through the wedges. Finally, a ruffled conical sleeve covers most of the actuator, except the top frame and motors, resembling the wrinkled skin of an elephant trunk. This sleeve restrains transverse deformations and increases the actuator's overall integrity.

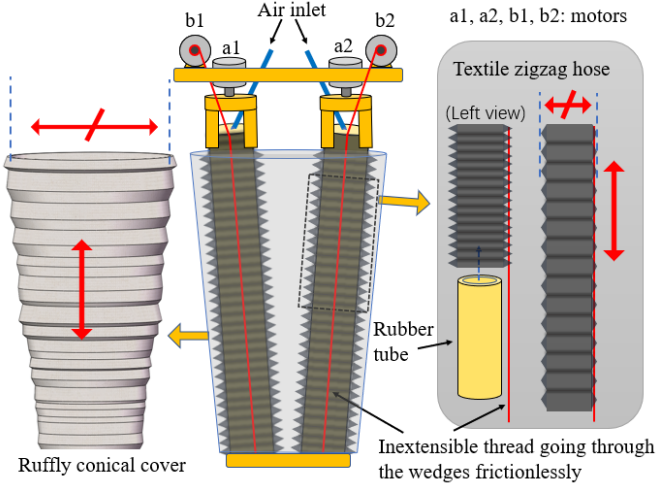

Figure 2. Mechanical design of the SEMI-TRUNK.

### B. Working principle and motion patterns.

A brief comparison and explanation of all six motion/pose patterns are given in Fig. 3. Detailed analyses are provided below.

In the following analysis, the pressures applied in the two tubes are all same and applied simultaneously. This restriction is intended to make the explanations clearer. Even with this restriction, six motion patterns can be realized; applying different pressures just makes them more complicated and adjustable.

To linearly extend the SEMI-TRUNK, we first control motors b1 and b2 to rotate in opposite directions so that the thread lengths increase by $l$ on both sides. We neglect the friction of the threads sliding inside the wedges of the hoses. Thus, when the same pressure $p$ is applied in both tubes, they perform linear extension from their initial length $L_0$ to $L$, the length after being pressurized, with $L$ satisfying the following relationship:

$$L < L_0 + l. \qquad (1)$$

Then, let us consider another situation. Suppose we keep the initial thread length $L_0$ and apply the same pressure $p$ in both tubes. In this case, not only the radial expansions of two tubes are blocked by the hose, the longitudinal extension on the front sides of the tubes are also restricted by the two threads. Thereby, the SEMI-TRUNK bends towards the top front, like a letter "C" in a vertical plane, as shown in the 3rd column of Fig. 3.

To realize J-shaped bending, we first control motors a1 and a2 to rotate the top parts of the tubes 90° in opposite directions. We define this rotation angle as the top twisting angle $θ$, which is positive when the tube is rotated to the left and negative to the right. In the following context, $θ$ of the left tube is denoted as $θ_1$ and $θ$ of the right tube is denoted as $θ_2$. In this case, $θ_1$ equals -90° and $θ_2$ equals 90°. The inextensible threads of the SEMI-TRUNK start from the front edges at the bottom and extend to the left and right edges at the top of the tubes.

Naturally, when the same pressure $p$ is applied in each tube, the tube tends to curve along the inextensible thread. Because the thread is twisted along the tube, if we decompose the curving force on every sectional plane along one tube, we can see that as the sectional plane shifts from the bottom to the top, the forward component of the curving force decreases from the maximum to 0; meanwhile the transverse component of the curving force increases from 0 to the maximum, as shown in the 4th column of Fig. 3.

At the same time, due to the existence of the ruffly conical cover, the leftward force component in the left tube and the rightward force component in the right tube always cancel each other out in the horizontal direction; then, only the forward curving forces remain and result in two-dimensional bending, with the curvature decreasing from the tip to the top of the actuator. When observed from the right side, this deformation is like a letter "J." Thus, we call this the J-shaped bending.

Based on the above analysis, we can also notice that if the top twisting angle $θ$ is larger than 90° (or less than -90°), top parts of the thread tend to twist to the back semi-circle of the tube so that this part of the actuator will generate a backward curving force after being pressurized. For instance, let us consider the situation when we twist the left tube -180° and the right tube 180° and then pressurize them with the same pressure $p$. Naturally, the threads in the lower half remain forward while in the top half, the threads turn backward. Therefore, in the lower half part of the SEMI-TRUNK, symmetric curving forces point to the right-front and left-front, and in the upper half part of the SEMI-TRUNK, a symmetric right-backward curving force and left-backward curving force

| Motion name | A) Linear extension | B) C-shaped bending | C) J-shaped bending | D) S-shaped bending | E) Helical bending | F) Spiral bending |
|---|---|---|---|---|---|---|
| Top section views<br>🟨 Extensible part<br>🟥 Inextensible part<br>← Curving direction | Back / Front | | Upper half / Lower half | Upper half / Lower half | | |
| Compositions before two chambers are pressurized | b1, b2 rotate opposite angles; Wire length increased length $l$ | $\theta_1 = \theta_2 = 0$ | $\theta_1 = -90°$, $\theta_2 = 90°$ | $\theta_1 = -180°$, $\theta_2 = 180°$ | $\theta_1 = \theta_2 = -90°$ | $\theta_1 = -90°$, $\theta_2 \in (0, 90°)$ |
| Expected motions after two chambers are pressurized | $p$, $p$; $L$ | $p$, $p$; $L_0$, $L$ | $p$, $p$ | $p$, $p$ | $p$, $p$ | $p$, $p$ |

Figure 3. Chart of the motion patterns to be generated by the SEMI-TRUNK.

exist. Like the J-shaped bending, all the symmetric transverse force components cancel each other out, and only the forward curving components in the lower half and the backward curving components in the upper half remain. The SEMI-TRUNK performs S-shaped bending, as shown in the 5th column of Fig. 3

To realize 2D deformation, we rotate two tubes towards opposite directions with the same angles. In contrast, to realize 3D deformation, an asymmetric structure is used.

Let us consider the situation when both tubes are rotated by the same angle to the left, with $\theta_1$ and $\theta_2$ equaling -90° for example, and then pressurized. Because the curving directions in the two tubes are identical, they do not cancel out but enhance each other. The whole actuator curves along the inextensible thread and makes a helical bending with the influence of gravity. When $\theta$ is negative, the SEMI-TRUNK makes a counter-clockwise helix, and with a positive $\theta$, the SEMI-TRUNK makes clockwise helix (top view), as shown in the 6th column of Fig. 3.

Spiral bending can be realized when the left tube is rotated -90° while the right tube is rotated with a positive angle less than 90° (typically 0-40°), with the same pressure applied. The two tubes tend to twist in opposite directions, but because the left tube generates a larger transverse curving component, the SEMI-TRUNK twists in a counter-clockwise direction. With the resistance force of the right tube, the curvature in the upper part of the SEMI-TRUNK is less than that of the helical bending when $\theta_1$ and $\theta_2$ equal -90°. In the lower part, the forward curving components get larger; meanwhile, the transverse components get smaller, considering the influence of gravity, and both tubes tend to curve towards the same direction, generally inside a 2D plane, as shown in the 7th column of Fig. 3.

In the 2nd row of Fig. 3, the top section views show the extensible and inextensible parts along the directions of the curving forces in two tubes. The 3rd row shows how the top twisting angles are changed and the corresponding twisted sub-actuators. In the final row, the expected deformations are shown after the SEMI-TRUNK is pressurized, where the black rings represent the transverse restrictions of the zigzag hoses and the red line represents the inextensible threads.

Note that there are many possible shapes in the spiral bending motion, and Fig. 3 only shows one pose of this motion pattern. An obvious factor that can largely enrich the deformation shapes is the inextensible thread length $L_0$. When a constant pressure is maintained, increasing or decreasing the threads' lengths together can change the curvatures in the "C," "J," and "S" deformations, but it cannot only change the height of the tip point while keeping the whole stiffness of the actuator in these deformations. In addition, applying different thread lengths in the helical or spiral deformation can largely increase the complexity with adjustable $\theta$ or $p$ at the same time.

III. MODELING

This section describes the models we made to quantitatively demonstrate the deformation processes of different motion patterns of the SEMI-TRUNK based on the working principles shown in section II. Because spiral bending has too many possible shapes, this motion pattern was not modeled. For the other motion patterns, mathematical models were established for the linear extension motion and the C-shaped bending motion. Simulation models based on the finite element method (FEM) were built for the C-shaped bending, J-shaped bending, S-shaped bending, and helical bending motions to reveal their morphological changes and the displacements of the tip point of the actuator under different pressures. Because the SEMI-TRUNK is designed to have a much larger length than its width to imitate the elephant trunk, we regard elongation of the SEMI-TRUNK as the same as the elongation of the sub-actuators in our modeling.

In the linear extension, after the two sub-actuators are pressurized with the same pressure $p$, considering the force balance at the bottom connector, we have

$$mg + 2p\frac{\pi d_i^2}{4} = 2EA\frac{L-L_0}{L_0}, \tag{2}$$

where $m$ is the mass of the actuation part of the SEMI-TRUNK without the motor frame, $g$ is the gravitational acceleration, $d_i$ is the inner diameter of the rubber tube, $E$ is the Young's modulus of the tube, and $A$ is the sectional area of the tube, which we assume as constant during the deformation. $L$ is the current length of the actuator after pressurization, and the length $A$ can be calculated using the following equation:

$$A = \frac{\pi(d_o^2 - d_i^2)}{4}, \tag{3}$$

where $d_o$ is the outer diameter of the tube. Using (2) and (3), we obtain the relationship between the pressurized actuator length and the input pressure as follows:

$$L = (\frac{2mg + p\pi d_i^2}{E\pi(d_o^2 - d_i^2)} + 1)L_0. \tag{4}$$

For the C-shaped bending, we ignore the influence of gravity, and thus the deformed shape of the actuator can be regarded as an arc with inner diameter $D$ and center angle $\beta$ inside a plane, as shown in Fig. 4(a).

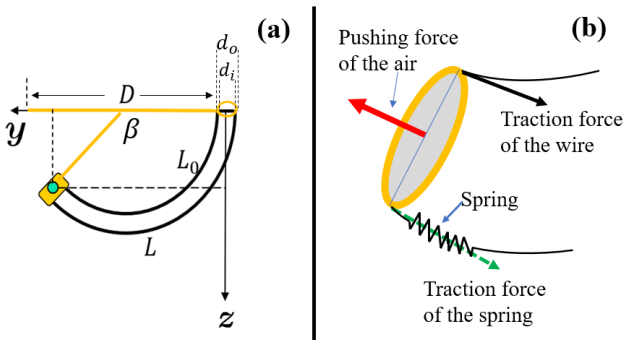

Figure 4. (a) Gravity is ignored to regard the C-shaped bending motion as an arc. (b) Force balance and torque at the tip point of each sub-actuator.

We build an x-y-z coordinate system with the origin set at the center of the bottom plane of the motor frame. Observed from the front, the x-axis points rightward, the y-axis points forward, and the z-axis points downward, as shown in Fig. 4(a). Because the whole actuator and inputs are symmetric about the y-z plane, we can analyze just one sub-actuator. We also regard the traction force of the whole rubber tube as a spring that lies along the outer edge of the tube, and assume the traction force of the spring after extension follows Hooke's law, as shown in Fig. 4(b).

In this case, if we consider the torque balance and force balance at the tip of each sub-actuator formed by the pushing force of the inside air, the traction forces of the spring and the thread shown in Fig. 4(b), we have

$$p\frac{\pi d_i^2}{4} = T + k(L - L_0), \tag{5}$$

$$T\frac{d_o}{2} - k(L - L_0)\frac{d_0}{2} = 0, \tag{6}$$

where $T$ is the traction force of the thread, $k$ is the spring constant, and $L$ is current length of the outer edge of the curved tube. From (5) and (6), we have

$$L - L_0 = p\frac{\pi d_i^2}{8k}. \tag{7}$$

From Fig. 4(a), we have

$$L - L_0 = \beta(\frac{D}{2} + d_o) - \beta\frac{D}{2} = \beta d_o. \tag{8}$$

Using (7) and (8), we obtain

$$\beta = \frac{p\pi d_i^2}{8kd_o}, \tag{9}$$

and thus,

$$D = \frac{2L_0}{\beta} = \frac{16kd_o L_0}{p\pi d_i^2}. \tag{10}$$

We can see that with $L_0$ settled, $\beta$ has a proportional relationship with the input pressure $p$; meanwhile, $D$ has an inverse proportional relationship with $p$.

If we define the coordinates of the center part of the tip as $(x, y, z)$, we have

$$x = 0, y = \frac{D + d_o}{2}(1 - \cos\beta), z = \frac{D + d_o}{2}\sin\beta. \tag{11}$$

Based on (9), (10), and (11), the coordinate changes of the tip of the SEMI-TRUNK during the C-shaped bending motion can be represented by the following equation with an unknown coefficient $k$:

$$x = 0,$$

$$y = f_1(k, p) = (\frac{8kd_o L_0}{p\pi d_i^2} + \frac{d_o}{2})(1 - \cos\frac{p\pi d_i^2}{8kd_o}), \tag{12}$$

$$z = f_2(k, p) = (\frac{8kd_o L_0}{p\pi d_i^2} + \frac{d_o}{2})\sin\frac{p\pi d_i^2}{8kd_o}.$$

The $y$ and $z$ coordinates are both functions of $k$ and $p$. The best value of $k$ in the model can be easily chosen through least squares fitting with several pairs of experimental data under the same pressures. The fitting process is described in section V.

We also build FEM models for the C-shaped bending and three other motion patterns: J-shaped bending, S-shaped bending, and helical bending. The models were built with Abaqus CAE 2020. Because the modeling of the textile zigzag hose and the ruffly conical cover is complicated and would dramatically increase the computing cost in the simulation, we neglected the conical cover and built the models based on the equivalent structure of each motion pattern, in which the transverse fiber rings represent the radial restrictions of the zigzag hoses and the longitudinal fibers represent the twisted inextensible threads.

As Fig. 5(a)-(d) shows, the four models have similar structures, except for the arrangements of the inextensible threads, which makes it easy for us to adjust them. Due to the absence of the conical cover, the transverse components of the curving forces in the J-shaped bending and S-shaped bending can no longer be restricted. Therefore, we exchanged the inextensible threads in the left tube and right tube so that they did not generate outward transverse bending components but rather inward components. We set up the frictionless contact of the surfaces of two tubes so that when they contact each other, the transverse components can be still neutralized, as shown in Fig. 5(b) and (c).

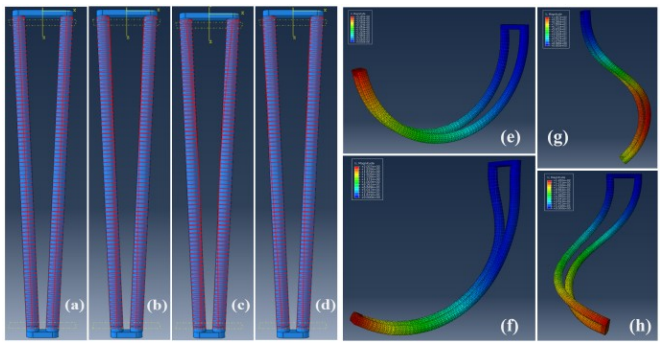

Figure 5. FEM models and simulation results: (a) C-shaped bending model, (b) J-shaped bending model (inextensible threads exchanged), (c) S-shaped bending model (inextensible threads exchanged), (d) helical bending model, (e) C-shaped bending result, (f) J-shaped bending model result, (g) S-shaped bending model result, and (h) helical bending result.

All the models have the same size and were used as a pattern to fabricate the SEMI-TRUNK prototype. The rubber tubes were 300 mm long, with a 7 mm inner diameter and 10 mm outer diameter and arranged in an inverted trapezoid shape. The top frame was 53 mm wide, and the bottom frame was 30 mm wide. The masses of the zigzag hose and the conical cover were added to the rubber tube, so the equivalent density of the rubber tube was larger than the actual density. The rubber tube was built using a hyper-elastic model and the threads were treated as a beam model. The stress-tension curve of the rubber material was tested and fitted using the Neo-Hookean model with $C10=0.46$ and $D1=0$, meaning the material is incompressible. The equivalent Young's modulus of the rubber we used in the mathematical model was 1.15 MPa.

During the simulations, the air pressure in the two tubes increased linearly from 0 to 0.2 MPa. The simulation results of four motion patterns are shown in Fig. 5(e)-(h). The displacements of different parts of the actuator are presented with different colors, with the unit of millimeters.

All the mathematical models and the simulation results were validated through experiments.

## IV. FABRICATION

In this section, we introduce the material selection and fabrication method of the SEMI-TRUNK. The materials we used are shown in Fig. 6(a). The rubber tubes had a 7 mm inner diameter and 10 mm outer diameter. The zigzag textile hose was cut from common garden watering hose, with a 10 mm inner diameter and 15 mm outer diameter. Both the hose and the tube had a 280 mm initial length, but the hose, possessing 78 wedges, could extend to at least 400 mm without external force. We used a sewing machine to make the textile conical cover from nylon cloth, whose smooth surface contributed to decreasing the friction between the actuator and the cover. The bottom connector was 3D-printed, and the center distance between cylindrical shafts was designed to be 15 mm to connect the two tubes closely.

We first inserted the two tubes into the zigzag hoses, sewed the threads through the wedges of corresponding hoses, and then connected them using the bottom connector. In this way, two sub-actuators were prepared as shown in Fig. 6(c). To decrease the friction between the thread and the hose, we passed the threads through the wedges once every four wedges.

The top parts of the sub-actuators were connected to a 3D-printed motor frame, whose structure is shown in Fig. 6(b). The two tubes were connected to two stepper motors through vertical rotational shafts with a 38 mm center-to-center distance. The inextensible threads were connected to two other stepper motors by pulleys so that when the motors produced rotation, the lengths of the inextensible threads could be adjusted. The stepper motors we used were 24BYJ-48, which could generate pull-in torque larger than 29.4 mN·m. We used the ULN2003 motor driver to drive them under 12 V DC, and all four motors were controlled using an Arduino Mega 2560 controller.

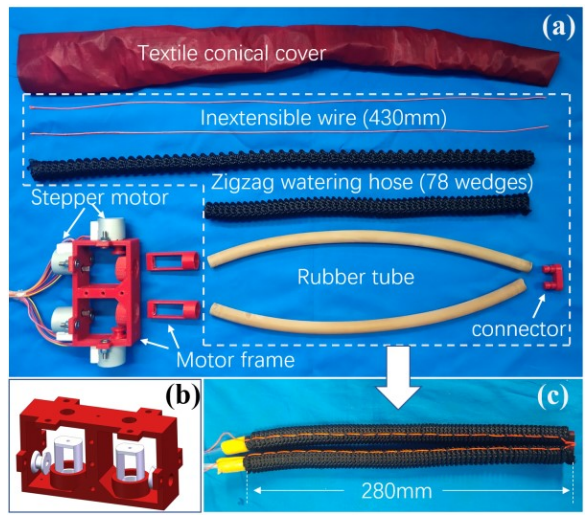

Figure 6. (a) Materials used to fabricate the SEMI-TRUNK. (b) Structure of the 3D-printed motor frame after assembly. (c) Two sub-actuators assembled from rubber tubes, zigzag-surface watering hoses, and inextensible threads.

After the sub-actuators were connected to the motor frame, the main part of the SEMI-TRUNK was finished as shown in Fig. 7(a). The length of the actuation part was kept at 300 mm.

The upper width of the actuation part was 53 mm, and the lower width was 30 mm. The dip angle of the sub-actuator $\varphi$, which is the closed angle between the sub-actuator and the bottom, was 87.648°. The initial length of the conical cover was 450 mm, but after we inserted the actuator, it was shortened to have the same length as the actuation part, thus becoming the ruffly conical cover that we needed.

Fig. 7(b) shows the size of the SEMI-TRUNK with the conical cover. The prototype was 370 mm long and weighed 245 g, but its actuation part only weighed 78 g. Its widest part reached 120 mm, and its narrowest part was only 30 mm wide.

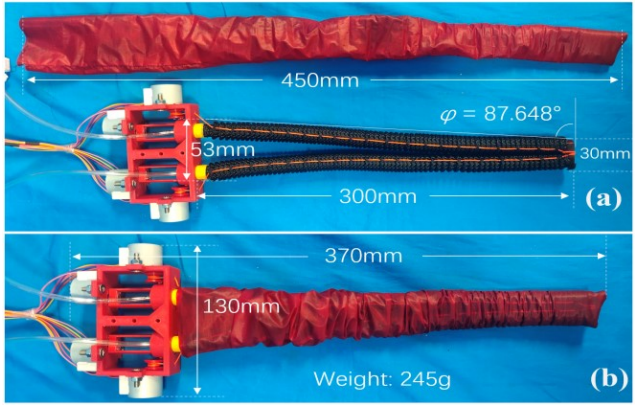

Figure 7. (a) SEMI-TRUNK without the conical cover. (b) SEMI-TRUNK covered with the ruffly conical cover.

V. EXPERIMENTS

This section explains how we designed and conducted experiments to evaluate the effectiveness of our mathematical models and FEM models.

For the linear extension motion, we used a ruler to measure the length change of the SEMI-TRUNK under linearly increased pressure from 0 to 0.2 MPa and then compared the result with that of the mathematical calculation using (4). The comparison result is shown in Fig. 8.

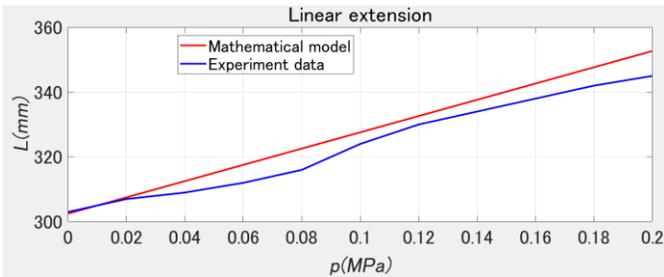

Figure 8. Comparison between the mathematical model and the experiment result on linear extension of the SEMI-TRUNK.

From Fig. 8, we can conclude that the mathematical model generally fits the experiment results with the prototype. We noticed that the initial lengths of the model and the prototype have almost the same value, while as the pressure increases, the actual length of the actuator remains smaller than the modeling result. We believe this error is due to the friction inside the prototype (rubber tubes and the zigzag hoses for instance), which we did not include in our model.

For the other four motion patterns, we used the experimental device shown in Fig. 9(a) to measure and record the coordinate changes of the tip of the SEMI-TRUNK. As the figure shows, a yellow marker was attached to the tip of the SEMI-TRUNK. On the right side of the actuator was a depth camera, an Intel Realsense D435. Possessing two parallelly arranged infrared cameras, the D435 can calculate the depth distance of any point inside its frame based on the differences of the pixel coordinates of the two cameras. The measurement error of the camera reaches 2% in the range from 0.2 m to 2 m.

As shown in Fig. 9(a), two coordinate systems were set. We defined the world coordinate system x-y-z with the origin set at the bottom center of the motor frame. The x-axis points right, the y-axis points forward, and the z-axis points downward. The origin of the camera coordinate system $x'-y'-z'$ lies at the inferred center point the two cameras, together with a backward x-axis, downward y-axis, and leftward z-axis.

We wrote Python programs to control the depth camera, making it able to distinguish yellow from other colors. After recognizing the yellow target, the camera could calculate the smallest rectangle to cover all the yellow part and return the depth distance of the center point of the rectangle. Then we could calculate the coordinates of the point in the camera coordinate system.

As the experiments began, we increased the air pressure in the two sub-actuators of SEMI-TRUNK simultaneously and linearly from 0 to 0.2 MPa, as in the FEM simulations. As Fig. 9(b) shows, the coordinates of the tip of the actuator in the camera coordinate system were printed on the captured frame and recorded. The video stream consisted of 30 frames per second with 640×480 pixels in each frame. After that, we designed a Gaussian filter to suppress the measurement error in the recorded data and then transform them into the coordinates in the world coordinate system.

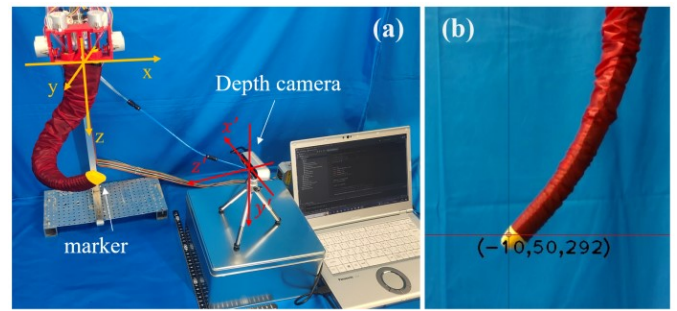

Figure 9. (a) Experimental setting. (b) Coordinates of the tip shown on the captured frame of the depth camera.

The yellow marker was 20 mm long, and thus the target point captured by the camera was assumed to be 290 mm from the bottom plane of the motor frame in the initial condition. For the FEM simulation model, we exported the coordinate change of the tip of the model from the initial coordinates (0, 0, 290).

To validate the mathematical model of the C-shaped bending, we first needed to choose an optimal value of $k$ in (12) using least squares fitting. As shown in Table I, four pairs of data were chosen and used in the least squares fitting after the experiment on C-shaped bending. Thus, the optimal value of k was calculated using (13).

$$k = \arg\min \sum_{1}^{4}(f_1(k,p_i)-y_i)^2 +(f_2(k,p_i)-z_i)^2. \quad (13)$$

Using (13) and the data in Table I, k=200.6 was chosen for the model (12) and applied in the following comparison.

TABLE I. FOUR PAIRS OF DATA USED TO OPTIMIZE $K$

| Air pressure p (MPa) | x coordinate | y coordinate | z coordinate |
|---|---|---|---|
| 0.05 | 0 | 80 | 276 |
| 0.10 | 0 | 125 | 239 |
| 0.15 | 0 | 157 | 185 |
| 0.20 | 0 | 171 | 138 |

Fig. 10(a) shows the comparison of the experiment result, mathematical model, and FEM model of the C-shaped bending. Fig. 10(b)-(d) shows the comparisons of the experiment results, mathematical calculations, and FEM simulations of the J-shaped bending, S-shaped bending, and helical bending, respectively.

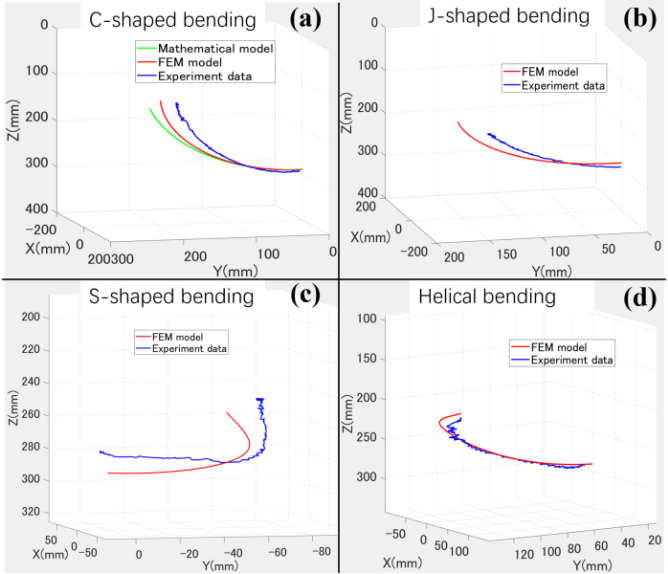

Figure 10. Comparison of the mathematical model, the FEM simulations, and the experiment results on the coordinate changes of the tip of the SEMI-TRUNK during (a) C-shaped bending, (b) J-shaped bending, (c) S-shaped bending, and (d) helical bending.

From Fig. 10, we can see that the general tendency of the real deformations of the SEMI-TRUNK fit with the corresponding mathematical and FEM models. Meanwhile, the trajectory cannot completely match the simulation results due to several reasons. The first and foremost reason is the friction forces inside the prototype actuator. As we controlled the motors to rotate the top part of the sub-actuators to switch from one motion pattern to another, the transverse friction between the zigzag hoses and the conical cover tended to prevent rotation, leading to an inhomogeneous slope of the inextensible thread, meaning that there are different fiber angles in different parts of one sub-actuator.

With different fiber angles in one tube, the sub-actuator may not deform exactly as we expect based on its working principles. Moreover, the asymmetric fiber angles of two sub-actuators may also appear due to the randomly distributed wrinkles on the conical cover.

Evidence of our claim is that Fig. 10(a)-(c) shows that the more we rotated the sub-actuators (from 0 to 180°), the larger the error becomes as the pressure increases. Because the more we rotate, the more the inner friction may impact the inextensible thread's slope, the error can be amplified by the pressurization process.

## VI. APPLICATION

Based on six different motion patterns and the large range of motions that the SEMI-TRUNK can achieve, we designed an application scenario to show the effectiveness of applying the SEMI-TRUNK to realize a complicated but very common task in the real world.

As shown in Fig. 11(a), a small plastic bottle of green tea and a cup were placed on a platform. The cylindrical bottle was 205 mm high, and its maximum diameter was 65 mm. The bottle weighed 18 g and the tea inside it weighed 42 g. Our target in this scenario was to control the SEMI-TRUNK to first grab this bottle of green tea, lift it up, and then pour it into a nearby cup. We chose this scenario because fetching a cup of tea satisfies the general expectation of what a practical robot arm should be able to do to assist humans in the real world. In addition, it is an appropriate test for evaluating the flexibility and practicability of the SEMI-TRUNK.

With the initial status shown in Fig. 11(a), the process of grabbing and pouring the tea proceeded as follows:

(1) Turn the top twisting angles $\theta_1$ and $\theta_2$ to -90°, and then pressurize the two tubes to 0.1 MPa. This step forms the helical bending motion pattern, but with a small pressure, the actuator is only lifted up a little, preparing it to approach the bottle from the left side, as shown in Fig. 11(b).

(2) Rotate $\theta_2$ to 10°. Due to this rotation, the tip of the actuator approaches the lower part of the bottle, as shown in Fig. 11(c).

(3) Increase pressures in both tubes to 0.2 MPa. Due to the different top twisting angles, spiral bending will appear. As shown in Fig. 11(d), the tip of the actuator embraces the bottle, preparing to grab it.

(4) Increase the pressure in the left tube to 0.25 MPa. With larger pressure, the inner diameter of the curved actuator decreases, the actuator grabs the bottle. The tension inside the left sub-actuator becomes larger to help the actuator make a counter-clockwise screwing motion, thus lifting the lower part of the bottle, as shown in Fig. 11(e).

(5) Decrease the pressure in the right tube to 0.1 MPa. Due to the larger pressure difference between the left and right tubes, the counter-clockwise screwing grows to a larger scale. Thus, the lower part of the bottle is further lifted up. As shown in Fig. 11(f), the bottle is now close to a horizontal posture, and the tea inside is positioned to be poured.

(6) Adjust the top twisting angles $\theta_1$ and $\theta_2$ together to rotate the actuator entirely until the mouth of the bottle is located above the cup, and then increase the pressure in the left tube to 0.3 MPa. Naturally, the lower part of the bottle is further lifted so that the tea inside the bottle is poured into the cup, as shown in Fig. 11(g). This completes the process.

The maximum payload of this scenario was about 68 g under 0.2 MPa, but this was slightly influenced by the thread length during the grabbing and transferring process. All the control signals were manually given from a laptop to the Arduino controller through a serial port to make the actuation speed adjustable. With sufficient practice, the whole grabbing-pouring procedure could be achieved within 120 s with a success rate of about 80%.

This application validated the effectiveness of the SEMI-TRUNK for completing complex real-world tasks.

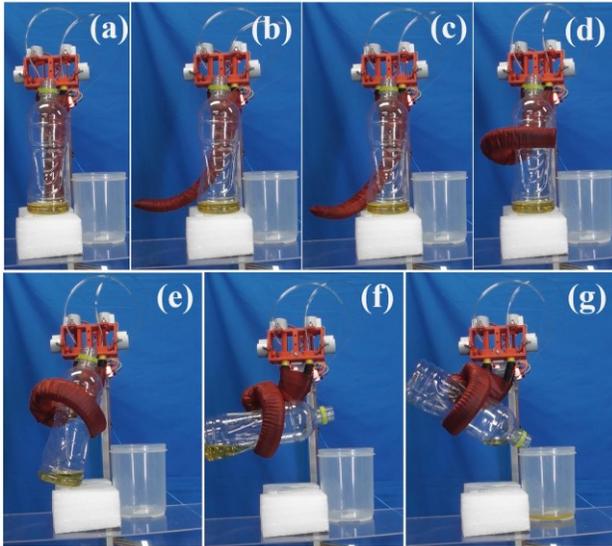

Figure 11. Application of grabbing and pouring a small bottle of green tea using the proposed SEMI-TRUNK.

VII. CONCLUSION

This paper proposes a new pneumatic soft actuator that can perform various bending and extending motions similar to those of an elephant's trunk by arranging two flexible tubes with a variable restraint function in parallel. Each flexible tube has a thread arranged in the axial direction, and has a variable restraint function that can adjust its length and inclination angle. It was confirmed that the composite wave of these two tubes could generate at least six different types of waveforms as superposition. Several mathematical models and FEM simulations were established to show its diversity in deforming and predict tip position changes during different deformations, which were validated through experiments. Furthermore, we confirmed that the prototype SEMI-TRUNK, which was developed based on our analyses, worked as intended, and that it could successfully grab a plastic bottle and pour tea into a cup. In the future, we are planning to pursue a method for shape feedback by installing a flexible sensor to detect the complex shape of the actuator.